\documentclass[11pt,letterpaper]{article}
\usepackage[letterpaper]{geometry}
\usepackage{acl2012}
\usepackage{times}
\usepackage{latexsym}
\usepackage{amsmath}
\usepackage{amssymb}
\usepackage{amsthm}
\usepackage{bbm}
\usepackage{mathtools}
\usepackage{multirow}
\usepackage{amsfonts}
\usepackage{url}
\makeatletter
\newcommand{\@BIBLABEL}{\@emptybiblabel}
\newcommand{\@emptybiblabel}[1]{}
\makeatother
\usepackage[hidelinks]{hyperref}

\newcommand{\RN}[1]{%
  \textup{\uppercase\expandafter{\romannumeral#1}}%
}
\theoremstyle{definition}
\newtheorem{definition}{Definition}[]
\usepackage[ruled,vlined,linesnumbered]{algorithm2e}
\usepackage[normalem]{ulem}
\usepackage{xcolor}

\newcommand{\com}[1]{}

\usepackage{color,colortbl}
\usepackage{lipsum}

\newcommand{\roi}[1]{}
\newcommand{\rotem}[1]{}

\title{Replicability Analysis for Natural Language Processing: Testing Significance with Multiple Datasets}

\author{Rotem Dror\\
\And Gili Baumer\\
\hspace{4cm} Faculty of Industrial Engineering and Management, Technion, IIT\\
\hspace{4cm}          {\tt \{rtmdrr@campus$|$sgbaumer@campus$|$marinabo$|$roiri\}.technion.ac.il}
\And Marina Bogomolov\\
\And Roi Reichart\\
}

\date{}

\begin{document}
\maketitle
\begin{abstract}
With the ever growing amounts of textual data from a large variety of languages, domains and genres, it has become standard to evaluate NLP algorithms on multiple datasets in order to ensure consistent performance across heterogeneous setups. However, such \textit{multiple comparisons} pose\rotem{transition - should this be pose or poses?} significant challenges to traditional statistical analysis methods in NLP and can lead to erroneous conclusions. 
In this paper we propose a \textit{Replicability Analysis} framework for a statistically sound analysis of multiple comparisons between algorithms for NLP tasks. We discuss the theoretical advantages of this framework over the current, statistically unjustified, practice in the NLP literature, and demonstrate its empirical value across four applications: multi-domain dependency parsing, multilingual POS tagging,  cross-domain sentiment classification and word similarity prediction. \footnote{Our code is at: https://github.com/rtmdrr/replicability-analysis-NLP .}
\end{abstract}

\section{Introduction}
\label{sec:intro}
	
The field of Natural Language Processing (NLP) is going through the data revolution. With the persistent increase of the heterogeneous web, for the first time in human history, written language from multiple languages, domains, and genres is now abundant. Naturally, the expectations from NLP algorithms also grow and evaluating a new algorithm on as many languages, domains, and genres as possible is becoming a de-facto standard.

For example, the phrase structure parsers of \newcite{Charniak:00} and \newcite{Collins:03} were mostly evaluated on the Wall Street Journal Penn Treebank \cite{Marcus:93}, consisting of written, edited English text of economic news. 
In contrast, modern dependency parsers are expected to excel on the 19 languages of the CoNLL 2006-2007 shared tasks on multilingual dependency parsing \cite{buchholz2006conll,nilsson2007conll}, and additional 
challenges, such as the shared task on parsing multiple English Web domains \cite{petrov2012overview}, are continuously  proposed.

Despite the growing number of evaluation tasks, the analysis toolbox employed by NLP researchers has remained quite stable. Indeed, in most experimental NLP papers, several algorithms are compared on a number of datasets where the performance of each algorithm 
is reported together with per-dataset statistical significance figures. 
However, with the growing number of evaluation datasets, it becomes more challenging to draw comprehensive conclusions from such comparisons. This is because although the probability of drawing an erroneous conclusion from a single comparison is small, with multiple comparisons the probability of making one or more false claims may be very high.

The goal of this paper is to provide the NLP community with a statistical analysis framework, which we term \textit{Replicability Analysis}, that will allow us to draw statistically sound conclusions in evaluation setups that involve multiple comparisons. The classical goal of replicability analysis is to examine the consistency of findings across studies in order to address the basic dogma of science, that a finding is more convincingly true if it is replicated in at least one more study \cite{heller2014deciding,patil2016statistical}. We adapt this goal to NLP, where we wish to ascertain the superiority of one algorithm over another across multiple datasets, which may come from different languages, domains and genres. Finding that one algorithm outperforms another across domains gives a sense of consistency to the results and a positive evidence that the better performance is not specific to a selected setup.\footnote{"Replicability" is sometimes referred to as "reproducibility". In recent NLP work the term reproducibility was used when trying to get identical results on the same data \cite{Neveol:16,Marrese-Taylor:17}. In this paper, we adopt the meaning of "replicability" and its distinction from "reproducibility" from \newcite{Peng:11} and \newcite{Leek:15} and refer to replicability analysis as the effort to show that a finding is consistent over different datasets from different domains or languages, and is not idiosyncratic to a specific scenario.}

In this work we address two questions: \textbf{(1)} \textit{Counting:} For how many datasets does a given algorithm outperform another? and \textbf{(2)} \textit{Identification:} What are these datasets?  

When comparing two algorithms on multiple datasets, NLP papers often answer informally the questions we address in this work. In some cases this is done without any statistical analysis, by simply declaring better performance of a given algorithm for datasets where its performance measure is better than that of another algorithm, and counting these datasets. In other cases answers are based on the p-values from statistical tests performed for each dataset: declaring
better performance for datasets with p-value below the significance level (e.g. 0.05) and counting these datasets. While it is clear that the first approach is not statistically valid, it seems that our community is not aware of the fact that the second approach, which may seem statistically sound, is not valid as well. This may lead to erroneous conclusions, which result in adopting new (probably complicated) algorithms, while they are not better than previous (probably more simple) ones. 

In this work, we demonstrate this problem and show that it becomes more severe as the number of evaluation sets grows, which seems to be the current trend in NLP.  We adopt a known general statistical methodology for addressing the counting (question (1)) and identification (question (2)) problems, by choosing the tests and procedures which are valid for situations encountered in NLP problems, and giving specific recommendations for such situations.

Particularly, we first demonstrate (Sec.~\ref{sec:pre})  that the current prominent approach in the NLP literature: identifying the datasets for which the difference between the performance of the algorithms reaches a predefined significance level according to some statistical significance test, does not guarantee to bound the probability to make at least one erroneous claim. Hence this approach is error-prone when the number of participating datasets is large.
We thus propose an alternative approach (Sec.~\ref{sec:rep_analysis}). For question (1), we adopt the approach of \newcite{benjamini2009selective} to replicability analysis of multiple studies, based on the partial conjunction framework of \newcite{benjamini2008screening}. This analysis comes with a guarantee that the probability of overestimating the true number of datasets with effect is upper bounded by a predefined constant. For question (2), we motivate a multiple testing procedure which guarantees that the probability of making at least one erroneous claim on the superiority of one algorithm over another
is upper bounded by a predefined constant. 
\com{In Section~\ref{sec:pre} we review the statistical background for the proposed framework, which includes hypothesis testing based on multiple datasets and the partial conjunction framework introduced in \cite{benjamini2008screening}. In Section~\ref{sec:rep_analysis} we then present two estimators, recommended for two different setups in NLP, for the number of domains on which one algorithm is superior to another, and a method for identifying these domains.}

In Sections~\ref{sec:toy} and ~\ref{sec:experiments} we demonstrate how to apply the proposed frameworks to two synthetic data toy examples and four NLP applications: multi-domain dependency parsing, multilingual POS tagging, cross-domain sentiment classification and word similarity prediction with word embedding models. 
Our results demonstrate that the current practice in NLP for addressing our questions is error-prone, and illustrate the differences between it and the proposed statistically sound approach.

We hope that this work will encourage our community to increase the number of standard evaluation setups per task when appropriate (e.g. including additional languages and domains), possibly paving the way to hundreds of comparisons per study. This is due to two main reasons. First, replicability analysis is a statistically sound framework that allows a researcher to safely draw valid conclusions with well defined statistical guarantees. Moreover, this framework provides a means of summarizing a large number of experiments with a handful of easily interpretable numbers (see, e.g., Table.~\ref{table:khats}). This allows researchers to report results over a large number of comparisons in a concise manner, delving into details of particular comparisons when necessary.

\section{Previous Work}
\label{sec:prev}

Our work recognizes the current trend in the NLP community where, for many tasks and applications, the number of evaluation datasets constantly increases. We believe this trend is inherent to language processing technology due to the multiplicity of languages and of linguistic genres and domains. In order to extend the reach of NLP algorithms, they have to be designed so that they can deal with many languages and with the various domains of each. Having a sound statistical framework that can deal with multiple comparisons is hence crucial for the field. 

This section is hence divided to two. We start by discussing representative examples for multiple comparisons in NLP,  focusing on evaluation across multiple languages and multiple domains. We then discuss existing analysis frameworks for multiple comparisons, both in the NLP and in the machine learning literatures, pointing to the need for establishing new standards for our community.

\paragraph {Multiple Comparisons in NLP}

Multiple comparisons of algorithms over datasets from different languages, domains and genres have become a de-facto standard in many areas of NLP. 
Here we survey a number of representative examples. A full list of NLP tasks is beyond the scope of this paper.

A common multilingual example is, naturally, machine translation, where it is customary to compare algorithms across a large number of source-target language pairs. This is done, for example, with the Europarl corpus consisting of 21 European languages  \cite{koehn2005europarl,koehn2007experiments} and with the datasets of the WMT workshop series with its multiple domains (e.g. news and biomedical in 2017), each consisting of several language pairs (7 and 14 respectively in 2017).\com{\footnote{http://www.statmt.org/wmt17/}} 

Multiple dataset comparisons are also abundant in domain adaptation work. Representative tasks include named entity recognition \cite{guo2009domain}, POS tagging \cite{daume2009frustratingly}, dependency parsing \cite{petrov2012overview}, word sense disambiguation \cite{chan2007domain} and sentiment classification \cite{blitzer2006domain,blitzer2007biographies}. 

More recently, with the emergence of crowdsourcing that makes data collection cheap and fast \cite{snow2008cheap}, an ever growing number of datasets is being created. This is particularly noticeable in lexical semantics tasks that have become central in NLP research due to the prominence of neural networks. For example, it is customary to compare word embedding models \cite{Mikolov:13,Pennington:14,OSeaghdha:Korhonen:14,Levy:14,schwartz-reichart-rappoport:2015:Conll} on multiple datasets where word pairs are scored according to the degree to which different semantic relations, such as similarity and association, hold between the members of the pair \cite{Finkelstein:01,Bruni:14,Silberer:14,Hill:15}. In some 
works (e.g. \cite{Baroni:14}) these embedding models are compared across a large number of simple tasks. 
\rotem{Another recent popular example is work on compositional distributional semantics and sentence embedding. For example, \newcite{Wieting:15} compare six sentence representation models on no less than 24 tasks.}

As discussed in Section \ref{sec:intro}, the outcomes of such comparisons are often summarized in a table that presents numerical performance values, usually accompanied with statistical significance figures and sometimes also with cross-comparison statistics such as average performance figures. Here, we analyze the conclusions that can be drawn from this information and suggest that with the growing number of comparisons, a more intricate analysis is required.

\com{The results from each dataset are usually displayed in a comparison table and the statistically significant differences are marked. However, the tests usually applied for statistical evaluation in NLP are compatible only for a comparison made on a single dataset. I.e., it is not statistically valid to count the number of significant results and report them, since the true amount of significant outcomes can be much smaller in reality.}


\paragraph{Existing Analysis Frameworks}

Machine learning work on multiple dataset comparisons dates back to
\newcite{dietterich1998approximate} who raised the question: "given two learning algorithms and datasets from several domains, which algorithm will produce more accurate classifiers when trained on examples from new domains?". 
The seminal work that proposed practical means for this problem is that of \newcite{demvsar2006statistical}. Given performance measures for two algorithms on multiple datasets, the authors test whether there is at least one dataset on which the difference between the algorithms is statistically significant.
For this goal they propose methods such as a paired t-test, a nonparametric sign-rank test and a wins/losses/ties count, all computed  across the results collected from all participating datasets.
In contrast, our goal is to count and identify the datasets for which one algorithm significantly outperforms the other, which provides more intricate information,
especially when the datasets come from different sources.

\com{However, in the presence of a large number of comparisons, the proposed methods are not adequate for answering questions such as on \textit{which} datasets the differences are significant or even \textit{how many} such datasets exist. 
Intuitively, this is because when a large number of comparisons is performed even rare events are likely to happen.}

In NLP, several studies addressed the problem of measuring the statistical significance of results on a single dataset (e.g. \cite{berg2012empirical,sogaard2013estimating,sogaard2014s}).
\newcite{sogaard2013estimating} is, to the best of our knowledge, the only work that addressed the statistical properties of evaluation with multiple datasets.
For this aim he modified the statistical tests proposed in \newcite{demvsar2006statistical} to use a Gumbel distribution assumption 
on the test statistics, which he considered to suit NLP better than the original Gaussian assumption. However, while this procedure aims to estimate the effect size across datasets,
it answers neither the counting nor the identification question of Section~\ref{sec:intro}.\roi{1. Not clear to me what is the "reduced error". 2. Is \newcite{sogaard2013estimating} the only work that addresses this question ? If so, we should say that, instead of saying he was the first.}\rotem{changed here - reduced error to effect size. Is this better?}


In the next section we provide the preliminary knowledge from the field of statistics that forms the basis for the proposed framework and then proceed with its description. 

\section{Preliminaries}
\label{sec:pre} 
We start by formulating a general hypothesis testing framework for a comparison between two algorithms. 
This is the common type of hypothesis testing framework applied in NLP, its detailed formulation will help us develop our ideas.
\subsection{Hypothesis Testing}
We wish to compare between two algorithms, $A$ and $B$. Let $X$ be a collection of datasets $X = \{X^1, X^2, \ldots, X^N\}$, where for all
$i \in \{1, \ldots ,N\}, X^i = \{x_{i,1},\ldots, x_{i,n_i}\}$ .
\com{
\[X =  \left\{ \begin{array}{ll}
     X^1= & (x_{1,1},\ldots, x_{1,n_1}),\\
     X^2= & (x_{2,1},\ldots, x_{2,n_2}), \\
     \vdotswithin{X^m} & \vdotswithin{(x_{m1},\ldots, x_{m,n_m}) } \\
     X^n= & (x_{n,1},\ldots, x_{n,n_n})\end{array}\right\}. \] 
}
Each dataset $X^i$ can be of a different language or a different domain. We denote by $x_{i,k}$ the granular unit on which results are being measured, which in most NLP tasks is a word or a sequence of words. The difference in performance between the two algorithms is measured using one or more of the evaluation measures in the set  
$\mathcal{M} = \{\mathcal{M}_1,\ldots,\mathcal{M}_m\}$.\footnote{To keep the discussion concise, throughout this paper we assume that only one evaluation measure is used. Our framework can be easily extended to deal with multiple measures.}

Let us denote $\mathcal{M}_j(ALG,X^i)$ as the value of the measure $\mathcal{M}_j$ when algorithm $ALG$ is applied on the dataset $X^i$. Without loss of generality, we assume that higher values of the measure are better.
We define the difference in performance between two algorithms, $A$ and $B$, according to the measure $\mathcal{M}_j$ on the dataset $X^i$ as:
\[\delta_j(X^i) = \mathcal{M}_j(A,X^i) - \mathcal{M}_j(B,X^i). \]

Finally, using this notation we formulate the following statistical hypothesis testing problem:
\begin{equation}
\label{eq:formulation}
\begin{split}
H_{0i}(j): & \delta_j(X^i) \le 0 \\
H_{1i}(j): & \delta_j(X^i) > 0.
\end{split}
\end{equation}

The null hypothesis, stating that there is no difference between the performance of algorithm $A$ and algorithm $B$, or that $B$ performs better, is tested versus the alternative statement that $A$ is superior. 
If the statistical test results in rejecting the null hypothesis, one concludes that $A$ outperforms $B$ in this setup.
Otherwise, there is not enough evidence in the data to make this conclusion. 

Rejection of the null hypothesis when it is true is termed \textit{type $\RN{1}$ error}, and non-rejection of the null hypothesis when the alternative is true is termed \textit{type $\RN{2}$ error}. 
The classical approach to hypothesis testing is to find a test that guarantees that the probability of making a type $\RN{1}$ error is upper bounded by a predefined constant $\alpha$, the test significance level, while achieving as low probability of type $\RN{2}$ error as possible, a.k.a achieving as high \textit{power} as possible. 

\com{Erroneous rejection of the null hypothesis when it is true is known as performing a type $\RN{1}$ error. The significance level of the test, denoted by $\alpha$, is the upper bound for the probability of making this type of error.

A statistical test is called \textit{valid} if it controls a certain type $\RN{1}$ error criterion, i.e., it guarantees to bound the error criterion, such as the significance level, by a predefined constant. However, one can achieve validity by never rejecting any null hypothesis, hence the quality of a statistical test is also being measured by its \textit{power}: the probability that it would reject a false-null hypothesis. 
In general, we wish to design tests that are both valid and powerful.}

We next turn to the case where the difference between two algorithms is tested across multiple datasets.

\subsection{The Multiplicity Problem} 
\label{subsec:adjmultiplicity}
Equation~\ref{eq:formulation} defines a multiple hypothesis testing problem when considering the formulation for all $N$ datasets. 
If $N$ is large, testing each hypothesis separately at the nominal significance level may result in a high number of erroneously rejected null hypotheses.\roi{This is the first place in the text where "nominal significance level" is used. This term has not been explained before.}
In our context, when the performance of algorithm $A$ is compared to that of algorithm $B$ across multiple datasets, and for each dataset algorithm $A$ is declared as superior based on a statistical test at the nominal significance level $\alpha$, the expected number of erroneous claims may grow as $N$ grows.

For example, if a single test is performed with a significance level of $\alpha = 0.05$, there is only a 5\% chance of incorrectly rejecting the null hypothesis. On the other hand, for 100 tests where all null hypotheses are true, the expected number of incorrect rejections is $100\cdot 0.05 = 5$. Denoting the total number of type $\RN{1}$ errors as $V$, we can see below that if the test statistics are independent then the probability of making at least one incorrect rejection is 0.994:
\begin{equation*}
\begin{split}
\mathbb{P}(V>0) = 1- \mathbb{P}(V=0) = &\\
1-\prod_{i=1}^{100}\mathbb{P}(\small{\text{no type $\RN{1}$ error in $i$}}) =& 1-(1-0.05)^{100}
\end{split}
\end{equation*}
This demonstrates that the naive method of counting the datasets for which significance was reached at the nominal level is error-prone.\roi{Again, I do not understand the usage of the word "may" in this context. I cannot map this sentence to a meaning.} Similar examples can be constructed for situations where some of the null hypotheses are false.

The multiple testing literature proposes various procedures for bounding the probability of making at least one type $\RN{1}$ error, as well as other, less restrictive error criteria (see a survey at \cite{farcomeni2007review}).
In this paper, we address the questions of counting and identifying the datasets for which algorithm $A$ outperforms $B$, with certain statistical guarantees regarding erroneous claims. While identifying the datasets gives more information when compared to just declaring their number, we consider these two questions separately. 
As our experiments show, according to the statistical analysis we propose the estimated number of datasets with effect (question 1) may be higher than the number of identified datasets (question 2).
We next present the fundamentals of the partial conjunction framework which is at the heart of our proposed methods.



\subsection{Partial Conjunction Hypotheses}  
We start by reformulating the set of hypothesis testing problems of Equation~\ref{eq:formulation} as a unified hypothesis testing problem. This problem aims to identify whether algorithm $A$ is superior to $B$ across all datasets. The notation for the null hypothesis in this problem is $H_{0}^{N/N}$ since we test if $N$ out of $N$ alternative hypotheses are true:
\begin{small}
\[H_{0}^{N/N}: \bigcup\limits_{i=1}^{N}H_{0i} \text{ is true} \quad vs. \quad H_{1}^{N/N}: \bigcap\limits_{i=1}^{N}H_{1i} \text{ is true}.\]
\end{small}

Requiring the rejection of the disjunction of all null hypotheses is often too restrictive for it involves observing a significant effect on all datasets, $i \in \{1,\ldots,N\}$. 
Instead, one can require a rejection of the \textit{global null hypothesis} stating that all individual null hypotheses are true, i.e., evidence
that at least one alternative hypothesis is true. This hypothesis testing problem is formulated as follows:
\[H_{0}^{1/N}: \bigcap\limits_{i=1}^{N}H_{0i} \text{ is true} \quad vs. \quad H_{1}^{1/N}: \bigcup\limits_{i=1}^{N}H_{1i} \text{ is true}.\]

Obviously, rejecting the global null may not provide enough information: it only indicates that algorithm $A$ outperforms $B$ on at least one dataset. Hence, this claim does not give any evidence for the consistency of the results across multiple datasets. 

A natural compromise between the above two formulations is to test the \textit{partial conjunction null}, which states that the number of false null hypotheses is lower than $u$, where $1\le u \le N$ is a pre-specified integer constant. The \textit{partial conjunction test} contrasts this statement with the alternative statement that at least $u$ out of the $N$ null hypotheses are false. 
\begin{definition}[\newcite{benjamini2008screening}]
\label{def:partial_conj}
Consider $N \ge 2$ null hypotheses: $H_{01},H_{02},\ldots,H_{0N}$, and let $p_1,\ldots,p_N$ be their associated $p-$values. Let $k$ be the true unknown number of false null hypotheses, then our question "Are at least $u$ out of $N$ null hypotheses false?" can be formulated as follows:
\[H_0^{u/N}:k<u \quad vs. \quad H_1^{u/N}:k\ge u .\]
\end{definition}

In our context, $k$ is the number of datasets where algorithm $A$ is truly better, and the partial conjunction test examines whether algorithm $A$ outperforms algorithm $B$ in at least $u$ of $N$ cases.

\roi{From here to the beginning of 4.1 quite a lot is changed, to avoid repititions. Please review the new text.}\newcite{benjamini2008screening} developed a general method for testing the above hypothesis for a given $u$. 
They also showed how to extend their method in order to answer our counting question. 
We next describe their framework and advocate a different, yet related method for dataset identification. 


\section{Replicability Analysis for NLP}
\label{sec:rep_analysis}
Referred to as the cornerstone of science \cite{moonesinghe2007most}, replicability analysis is of predominant importance in many scientific fields including psychology \cite{open2012open}, genomics \cite{heller2014deciding}, economics \cite{herndon2014does} and medicine \cite{begley2012drug}, among others. 
Findings are usually considered as replicated if they are obtained in two or more studies that differ from each other in some aspects (e.g. language, domain or genre in NLP).

The replicability analysis framework we employ \cite{benjamini2008screening,benjamini2009selective} is based on partial conjunction testing. Particularly, these authors have shown that a lower bound on the number of false null hypotheses with a confidence level of $1-\alpha$ can be obtained by finding the largest $u$ for which we can reject the partial conjunction null hypothesis $H_{0}^{u/N}$ along with $H_{0}^{1/N},\ldots, H_{0}^{(u-1)/N}$ at a significance level $\alpha.$
This is since rejecting $H_0^{u/N}$ means that we see evidence that in at least $u$ out of $N$ datasets algorithm $A$ is superior to $B$. This lower bound on $k$ is taken as our answer to the \textit{Counting} question of Section~\ref{sec:intro}.

In line with the hypothesis testing framework of Section~\ref{sec:pre}, the partial conjunction null, $H_0^{u/N}$, is rejected at level $\alpha$ if $p^{u/N} \le \alpha$, where $p^{u/N}$ is the partial conjunction $p$-value. Based on the known methods for testing the global null hypothesis (see, e.g., \cite{loughin2004systematic}), \newcite{benjamini2008screening} proposed methods for combining the $p-$values $p_1,\ldots,p_N$ of $H_{01},H_{02},\ldots,H_{0N}$ in order to obtain $p^{u/N}$.
Below, we describe two such methods and their properties. 

\subsection{The Partial Conjunction \texorpdfstring{$p-$}{p-}value}
\label{subsec:combinations}

The methods we focus at were developed in  \newcite{benjamini2008screening}, and are based on Fisher's and Bonferroni's methods for testing the global null hypothesis. For brevity, we name them \emph{Bonferroni} and \emph{Fisher}. We choose them because they are valid in different setups that are frequently encountered in NLP (Section~\ref{sec:experiments}):
Bonferroni for dependent datasets and both Fisher and Bonferroni for independent datasets.\footnote{For simplicity we refer to dependent/independent datasets as those for which the test statistics are dependent/independent. We assume the test statistics are independent if the corresponding datasets do not have mutual samples, and one dataset is not a transformation of the other.}

Bonferroni's method does not make any assumptions about the dependencies between the participating datasets and it is hence applicable in NLP tasks, since in NLP it is most often hard to determine the type of dependence between the datasets.
Fisher's method, while assuming independence across the participating datasets, is often more powerful than Bonferroni's method\com{or other methods which make the same independence assumption} (see \cite{loughin2004systematic,benjamini2008screening} for other methods and a comparison between them). Our recommendation is hence to use the Bonferroni's method when the datasets are dependent and to use the more powerful Fisher's method when the datasets are independent. 

Let $p_{(i)}$ be the $i$-th smallest $p-$value among  $p_1,\ldots,p_N$. The partial conjunction $p-$values are: 
\begin{align}
p^{u/N}_ {Bonferroni} = (N-u+1)p_{(u)} \qquad \qquad \quad \label{eq:1} \\
p^{u/N}_ {Fisher} = \mathbb{P}\left(\chi^2_{2(N-u+1)}\ge -2\sum_{i=u}^{N}\ln p_{(i)} \right) \label{eq:2}
\end{align}
where $\chi^2_{2(N-u+1)}$ denotes a chi-squared random variable with $2(N-u+1)$ degrees of freedom. 

To understand the reasoning behind these methods, let us consider first the above $p-$values for testing the global null, i.e., for the case of $u=1$. 
Rejecting the global null hypothesis requires evidence that at least one null hypothesis is false. Intuitively, we would like to see one or more small $p-$values. 

Both of the methods above agree with this intuition. Bonferroni's method rejects the global null if $p_{(1)}\leq \alpha/N$, i.e. if the minimum $p-$value is small enough, where the threshold guarantees that the significance level of the test is $\alpha$ for any dependency among the $p-$values $p_1,\ldots,p_N$. Fisher's method rejects the global null for large values of $-2\sum_{i=1}^{N}\ln p_{(i)}$, or equivalently for small values of $\prod_{i=1}^N p_i$. 
That is, while both these methods are intuitive, they are different. Fisher's method requires a small enough product of $p-$values as evidence that at least one null hypothesis is false. Bonferroni's method, on the other hand, requires as evidence at least one small enough $p-$value.
\com{
Now let us consider testing the partial conjunction hypothesis for $u>1$. If the alternative is true, i.e., at least $u$ null hypotheses are false, then one will find one or more false null hypotheses in any subset of $n-u+1$ hypotheses. 
Thus in order to see evidence that the alternative is true, it is intuitive to require rejection of the global null (the intersection hypothesis) for all subsets of $n-u+1$ hypotheses. Both Bonferroni's and Fisher's methods agree with this intuition. 
The $p-$value in Equation~\ref{eq:1} is below $\alpha$ if for every subset of $n-u+1$ hypotheses the Bonferroni's global null $p-$value is below $\alpha$ hence the global null is rejected at the significance level of $\alpha$ for every subset of $n-u+1$ hypotheses. 
Similarly, Fisher's method rejects $H_0^{u/n}$ if for all subsets of $n-u+1$ null hypotheses, the Fisher's global null $p-$value is below $\alpha$.}

For the case $u=N$, i.e., when the alternative states that all null hypotheses are false, both methods require that the maximal $p-$value is small enough for rejection of $H_0^{N/N}$. This is also intuitive because we expect that all the $p-$values will be small when all the null hypotheses are false. For other cases, where $1<u<N$, the reasoning is more complicated and is beyond the scope of this paper.

The partial conjunction test for a specific $u$ answers the question "Does algorithm A perform better than B on at least $u$ datasets?"
The next step is the estimation of the number of datasets for which algorithm $A$ performs better than $B$. 

\subsection{Dataset Counting (Question 1)}
\label{subsec:kestimator}
Recall that the number of datasets where algorithm $A$ outperforms algorithm $B$ (denoted with $k$ in Definition~\ref{def:partial_conj}) is the true number of false null hypotheses in our problem. \newcite{benjamini2008screening} proposed to estimate $k$ to be the largest $u$ for which $H_{0}^{u/N},$ along with $H_0^{1/N},\ldots,H_0^{(u-1)/N}$ is rejected. 
Specifically, the estimator $\hat{k}$ is defined as follows:
\begin{equation}
\hat{k} = \max\{u:p^{u/N}_*\le \alpha\},
\end{equation}
where $p^{u/N}_* = \max\{p^{(u-1)/N}_* ,p^{u/N} \}$, $p^{1/N}=p^{1/N}_*$
and $\alpha$ is the desired upper bound on the probability to overestimate the true $k$.\roi{We do not give any explanation as to why they use this non-standard formulation.}\rotem{the only explanation we have written before is that using p* instead of p will guarantee that $\mathbb{P}(\hat{k}>k) \le \alpha$, I don't know if we want to write this down.} It is guaranteed that $\mathbb{P}(\hat{k}>k) \le \alpha$ as long as the $p-$value combination method used for constructing $p^{u/N}$ is valid for the given dependency across the test statistics.\footnote{This result is a special case of Theorem 4 in \cite{benjamini2008screening}.} 
When $\hat{k}$ is based on $p^{u/N}_{Bonferroni}$ it is denoted with $\hat{k}_{Bonferroni}$, while when it is based on $p^{u/N}_{Fisher}$ it is denoted with  $\hat{k}_{Fisher}$.


A crucial practical consideration when choosing between $\hat{k}_ {Bonferroni}$ and $\hat{k}_ {Fisher}$ is the assumed dependency between the datasets.
As discussed in Section~\ref{subsec:combinations}, $p^{u/N}_{Fisher}$ is recommended when the participating datasets are assumed to be independent, while when this assumption cannot be made only $p^{u/N}_{Bonferroni}$ is appropriate. As the $\hat{k}$ estimators are based on 
the respective $p^{u/N}$s, the same considerations hold when choosing between them.

With the $\hat{k}$ estimators, one can answer the counting question of Section~\ref{sec:intro}, reporting that algorithm $A$ is better than algorithm $B$ in at least $\hat{k}$ out of $N$ datasets with a confidence level of $1-\alpha$. 
Regarding the identification question, a natural approach would be to declare the $\hat{k}$ datasets with the smallest $p-$values as those for which the effect holds. However, with $\hat{k}_{Fisher}$ this approach does not guarantee control over type $\RN{1}$ errors. In contrast, for $\hat{k}_{Bonferroni}$ the above approach comes with such guarantees, as described in the next section. 


\subsection{Dataset Identification (Question 2)}
\label{subsec:identification}
As demonstrated in Section~\ref{subsec:adjmultiplicity}, identifying the datasets with $p-$value below the nominal significance level and declaring them as those where algorithm $A$ is better than $B$ may lead to a very high number of erroneous claims. A variety of methods exist for addressing this problem. 
A classical and very simple method for addressing this problem is named the Bonferroni's procedure, which compensates for the increased probability of making at least one type $\RN{1}$ error by testing each individual hypothesis at a significance level of $\alpha' = \alpha /N$, where $\alpha$ is the predefined bound on this probability and $N$ is the number of hypotheses tested.\footnote{Bonferroni's correction is based on similar considerations as  $p^{u/N}_ {Bonferroni}$ for $u=1$ (Eq. 2). The partial conjunction framework (Sec.~\ref{subsec:combinations}) extends this idea for other values of $u$.} 
While Bonferroni's procedure is valid for any dependency among the $p-$values, the probability of detecting a true effect using this procedure is often very low, because of its strict $p-$value threshold.

%

Many other procedures controlling the above or other error criteria and having less strict $p-$value thresholds have been proposed. 
Below we advocate one of these methods: the \textit{Holm procedure} \cite{Holm:79}. This is a simple $p-$value based procedure that is concordant with the partial conjunction analysis when $p^{u/N}_{Bonferroni}$ is used in that analysis. Importantly for NLP applications,  Holm controls the probability of making at least one type $\RN{1}$ error for any type of dependency between the participating datasets (see a demonstration in Section~\ref{sec:experiments}).

Let $\alpha$ be the desired upper bound on the probability that at least one false rejection occurs, let $p_{(1)} \le p_{(2)} \le \ldots \le p_{(N)}$ be the ordered $p-$values and let the associated hypotheses be $H_{{(1)}}\ldots H_{{(N)}}$. The Holm procedure for identifying the datasets with a significant effect is given in below. 

\SetNlSty{}{}{)}
\begin{procedure}
\label{alg:holm}
\caption{Holm()}
Let $k$ be the minimal index such that $p_{{(k)}}>{\frac{\alpha }{N+1-k}}$.\\
Reject the null hypotheses $H_{{(1)}}\ldots H_{{(k-1)}}$ and do not reject $H_{{(k)}}\ldots H_{{(N)}}$.
If no such $k$ exists, then reject all null hypotheses.
\end{procedure}

The output of the Holm procedure is a rejection list of null hypotheses, 
the corresponding datasets are those we return in response to the identification question of Section~\ref{sec:intro}.
Note that the Holm procedure rejects a subset of hypotheses with p-value below $\alpha$. Each p-value is compared to a threshold which is smaller or equal to $\alpha$ and depends on the number of evaluation datasets $N.$ The dependence of the thresholds on $N$ can be intuitively explained as follows. The probability of making one or more erroneous claims may increase with $N,$ as demonstrated in Section \ref{subsec:adjmultiplicity}. Therefore, in order to bound this probability by a pre-specified level $\alpha,$ the thresholds for p-values should depend on $N.$

It can be shown that the Holm procedure at level $\alpha$ always rejects the $\hat{k}_{Bonferroni}$ hypotheses with the smallest $p-$values, where $\hat{k}_{Bonferroni}$ is the lower bound for $k$ with a confidence level of $1-\alpha$. Therefore, $\hat{k}_{Bonferroni}$ corresponding to a confidence level of $1-\alpha$ is always smaller or equal to the number of datasets for which the difference between the compared algorithms is significant at level $\alpha$. This is not surprising in view of the fact that without making any assumptions on the dependencies among the datasets, $\hat{k}_{Bonferroni}$ guarantees that the probability of making a too optimistic claim ($\hat{k} > k$) is bounded by $\alpha$, while when simply counting the number of datasets with p-value below $\alpha$, the probability of making a too optimistic claim may be close to 1, as demonstrated in Section \ref{sec:toy}.

\textbf{Framework Summary} Following Section~\ref{subsec:kestimator} we suggest to answer the counting question of Section~\ref{sec:intro} by reporting either $\hat{k}_{Fisher}$ (when all datasets can be assumed to be independent) or $\hat{k}_{Bonferroni}$ (when such an independence assumption cannot be made). Based on Section~\ref{subsec:identification} we suggest to answer the identification question of Section~\ref{sec:intro} by reporting the rejection list returned by the Holm procedure. 

Our proposed framework is based on certain assumptions regarding the experiments conducted in NLP setups. The most prominent of these assumptions states that for dependent datasets the type of dependency cannot be determined. Indeed, to the best of our knowledge, the nature of the dependency between dependent test sets in NLP work has not been analyzed before. In Section \ref{sec:conc} we revisit our assumptions and point on alternative methods for answering our questions. These methods may be appropriate under other assumptions that may become relevant in future. 

We next demonstrate the value of the proposed replicability analysis through toy examples with synthetic data (Section \ref{sec:toy}) as well as analysis of state-of-the-art algorithms for four major NLP applications (Section \ref{sec:experiments}). Our point of reference is the standard, yet statistically unjustified, counting method that sets its estimator, $\hat{k}_{count}$, to the number of datasets for which the difference between the compared algorithms is significant with $p-$value$ \le \alpha$ (i.e. $\hat{k}_{count}=\#\{i:p_i \le \alpha\}$).\footnote{We use $\alpha$ in two different contexts: the significance level of an individual test and the bound on the probability to overestimate $k$. This is the standard notation in the statistical literature.} 

\section{Toy Examples}
\label{sec:toy}

\begin{figure}
  \centering
  \includegraphics[width=\columnwidth, height=.18\textheight]{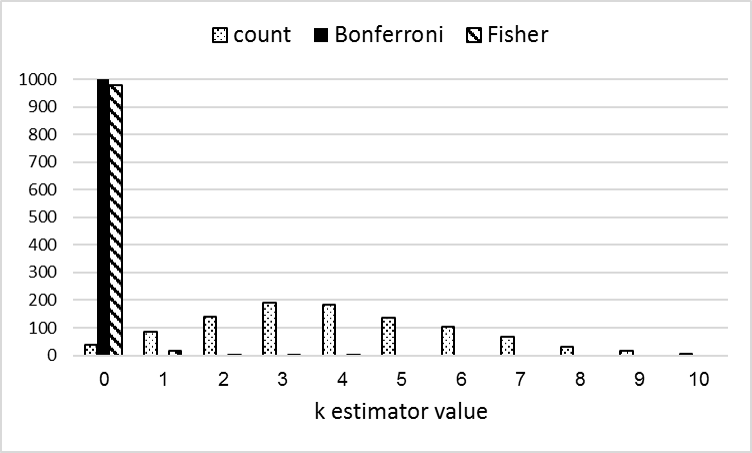}
  \caption{$\hat{k}$ histogram for the independent datasets simulation.}
  \label{fig:sim}
\end{figure}

For the examples of this section we synthesize $p-$values to emulate a test with $N =100$ hypotheses (domains), and set $\alpha$ to 0.05. \rotem{changed here - added footnote} We start with a simulation of a scenario where algorithm $A$ is equivalent to $B$ for each domain, and the datasets representing these domains are independent. We sample the 100 $p-$values from a standard uniform distribution, which is the $p-$value distribution under the null hypothesis, repeating the simulation  1000 times.

Since all the null hypotheses are true then $k$, the number of false null hypotheses, is $0$. 
Figure~\ref{fig:sim} presents the histogram of $\hat{k}$ values from all 1000 iterations according to $\hat{k}_{Bonferroni}$, $\hat{k}_{Fisher}$ and $\hat{k}_{count}$.

The figure clearly demonstrates that $\hat{k}_{count}$ provides an overestimation of $k$ while $\hat{k}_{Bonferroni}$ and $\hat{k}_{Fisher}$ do much better. Indeed, the histogram yields the following probability estimates: 
$\hat{P}(\hat{k}_{count}>k)=0.963$, $\hat{P}(\hat{k}_{Bonferroni}>k)=0.001$ and $\hat{P}(\hat{k}_{Fisher}>k)=0.021$ (only the latter two are lower than 0.05). This simulation strongly supports the theoretical results of Section \ref{subsec:kestimator}.\roi{Notice that for $\hat{k}_{count}$ $\alpha$ is used twice - once at the level of the individual dataset (i.e. only datasets with p-value lower than $\alpha$ are counted) and once as a confidence level for the counting (i.e. we would like the probability that $\hat{k}_{count} > k$ to be lower than $\alpha$ . Should we say something about this ? I am quite perplexed by this issue.} 

To consider a scenario where a dependency between the participating datasets does exist, we consider a second toy example. In this example we generate $N=100$ $p-$values corresponding to 34 independent normal test statistics, and two other groups of 33 positively correlated normal test statistics with $\rho=0.2$ and $\rho=0.5$, respectively. 
We again assume that all null hypotheses are true and thus all the $p-$values are distributed uniformly, repeating the simulation 1000 times. To generate positively dependent $p-$values we followed the process described in Section 6.1 of \newcite{benjamini2006adaptive}.\roi{There are a couple of details that I do not understand here due to lack of statistical knowledge. For example, how can the p-values be evenly distributed but still have positive correlations. Or, which type of correlation test does the $\rho$ refer to and so on. I wonder if we want to get into this. }

We estimate the probability that $\hat{k}>k=0$ for the three $\hat{k}$ estimators based on the 1000 repetitions and get the values of:
$\hat{P}(\hat{k}_{count}>k)=0.943$, $\hat{P}(\hat{k}_{Bonferroni}>k)=0.046$ and $\hat{P}(\hat{k}_{Fisher}>k)=0.234$. This simulation demonstrates the importance of using Bonferroni's method rather than Fisher's method when the datasets are dependent, even if some of the datasets are independent.

\section{NLP Applications}
\label{sec:experiments}

In this section we demonstrate the potential impact of replicability analysis on the way experimental results are analyzed in NLP setups. We explore four NLP applications: (a) two where the datasets are independent: multi-domain dependency parsing and multilingual POS tagging; and (b) two where dependency between the datasets does exist: cross-domain sentiment classification and word similarity prediction with word embedding models.

\subsection{Data}
\paragraph{Dependency Parsing} We consider a multi-domain setup, analyzing the results reported in \newcite{choi2015depends}. 
The authors compared ten state-of-the-art parsers from which we pick three: (a) Mate \cite{bohnet2010very}\footnote{\url{code.google.com/p/mate-tools}.} that performed best on the majority of datasets; (b) Redshift \cite{honnibal2013non}\footnote{\url{github.com/syllog1sm/Redshift}.} which demonstrated comparable, still somewhat lower, performance compared to Mate; and (c) SpaCy \cite{honnibal-johnson:2015:EMNLP}\footnote{\url{honnibal.github.io/spaCy}.} that was substantially outperformed by Mate.

All parsers were trained and tested on the English portion of the OntoNotes 5 corpus \cite{weischedel2011ontonotes,pradhan2013towards}, a large multi-genre corpus consisting of the following 7 genres: broadcasting conversations (BC), broadcasting news (BN), news magazine (MZ), newswire (NW), pivot text (PT), telephone conversations (TC) and web text (WB). Train and test set size (in sentences) range from 6672 to 34492 and from 280 to 2327, respectively (see Table 1 of \cite{choi2015depends}).
We copy the test set UAS results of \newcite{choi2015depends} and 
compute $p-$values using the data downloaded from \url{http://amandastent.com/dependable/}.

\com{For the multilingual setup we experiment with the TurboParser  \cite{martins2010turbo} on all 19 languages of the CoNLL 2006 and 2007 shared tasks on multilingual dependency parsing \cite{buchholz2006conll,nilsson2007conll}. We compare the performance of the first order parser trained with the MIRA algorithm to the same parser when trained with the perceptron algorithm, both implemented within the TurboParser.}

\paragraph{POS Tagging} 

We consider a multilingual setup, analyzing the results reported in \cite{Pinter:17}. The authors compare their \textsc{Mimick} model with the model of \newcite{Ling:15}, denoted with \textsc{char$\to$tag}. Evaluation is performed on 23 of the 44 languages shared by the Polyglot word embedding dataset \cite{AlRfou:13} and the universal dependencies (UD) dataset \cite{DeMarneffe:14}. \newcite{Pinter:17} choose their languages so that they reflect a variety of typological, and particularly morphological, properties. The training/test split is the standard UD split. We copy the word level accuracy figures of \cite{Pinter:17} for the low resource training set setup, the focus setup of that paper. The authors kindly sent us their $p$-values.

\paragraph{Sentiment Classification} 
In this task, an algorithm is trained on reviews from one domain and should classify the sentiment of reviews from another domain to the \textit{positive} and \textit{negative} classes.
For replicability analysis we explore the results of \newcite{ziser2016neural} for the cross-domain sentiment classification task of \newcite{blitzer2007biographies}. The data in this task consists of Amazon product reviews from 4 domains: books (B), DVDs (D), electronic items (E), and kitchen appliances (K), for the total of 12 domain pairs, each domain having a 2000 review test set.\footnote{\url{http://www.cs.jhu.edu/~mdredze/datasets/sentiment/index2.htm}} \newcite{ziser2016neural} compared the accuracy of their AE-SCL-SR model to MSDA \cite{chen2011automatic}, a well known domain adaptation method, and kindly sent us the required $p$-values.


\paragraph{Word Similarity}
We compare two state-of-the-art word embedding collections: (a) word2vec CBOW \cite{Mikolov:13} vectors, generated by the model titled the best "predict" model in \newcite{Baroni:14};\footnote{\url{http://clic.cimec.unitn.it/composes/semantic-vectors.html}. Parameters: 5-word context window, 10 negative samples, subsampling, 400 dimensions.} and (b) Glove \cite{Pennington:14} vectors generated by a model trained on a 42B token common web crawl.\footnote{http://nlp.stanford.edu/projects/glove/. 300 dimensions.} We employed the demo of \newcite{faruqui-2014:SystemDemo} to perform Spearman correlation evaluation of these vector collections on 12 English word pair datasets: WS-353 \cite{finkelstein2001placing}, WS-353-SIM \cite{agirre2009study}, WS-353-REL \cite{agirre2009study}, MC-30 \cite{miller1991contextual}, RG-65 \cite{rubenstein1965contextual}, Rare-Word \cite{Luong-etal:conll13:morpho}, MEN \cite{bruni2012distributional}, MTurk-287 \cite{radinsky2011word}, MTurk-771 \cite{halawi2012large}, YP-130 \cite{yang2006verb}, SimLex-999 \cite{hill2016simlex}, and Verb-143 \cite{baker2014unsupervised}.

\subsection{Statistical Significance Tests}
\label{subsec:methods}
We first calculate the $p-$values for each task and dataset according to the principals of $p-$values computation for NLP discussed in~\newcite{yeh2000more}, \newcite{berg2012empirical} and \newcite{sogaard2014s}.

For dependency parsing, we employ the a-parametric paired bootstrap test \cite{efron1994introduction} that does not assume any distribution on the test statistics. We choose this test because the distribution of the values for the measures commonly applied in this task is unknown.
We implemented the test as in \cite{berg2012empirical} with a bootstrap size of 500 and with $10^5$ repetitions.

For multilingual POS tagging we employ the Wilcoxon signed-rank test \cite{Wilcoxon:45}. The test is employed for the sentence level accuracy scores of the two compared models. This test is a non-parametric test for difference in measures, which tests the null hyphotesis that the difference has a symmetric distribution around zero. It is appropriate for tasks with paired continuous measures for each observation, which is the case when comparing sentence level accuracies. 

For sentiment classification we employ the McNemar test for paired nominal data \cite{mcnemar1947note}. This test is appropriate for binary classification tasks and since we compare the results of the algorithms when applied on the same datasets, we employ its paired version.
Finally, for word similarity with its Spearman correlation evaluation, we choose the Steiger test \cite{steiger1980tests} for comparing elements in a correlation matrix.


We consider the case of $\alpha = 0.05$ for all four applications. For the dependent datasets experiments (sentiment classification and word similarity prediction) with their generally lower $p-$values (see below), we also consider the case where $\alpha=0.01$.


\subsection{Results}
\label{subsec:res}
\begin{table}[b!]
\centering
\begin{tabular}{ | c | c | c | c | }
\hline
 & $\hat{k}_{count}$ & $\hat{k}_{Bonf.}$ & $\hat{k}_{Fish.}$ \\ \hline
\multicolumn{4}{|c|}{\textbf{Independent Datasets}} \\ \hline
\multicolumn{4}{|c|}{\textit{Dependency Parsing (7 datasets)}} \\ \hline
Mate-SpaCy & 7 & 7 & \textbf{7} \\ \hline
Mate-Redshift & 2 & 1 & \textbf{5} \\  \hline
%
\multicolumn{4}{|c|}{\textit{Multilingual POS Tagging (23 datasets)}} \\ \hline 
\textsc{Mimick}-Char$\rightarrow$Tag & 11 & 6 & \textbf{16} \\ \hline
\multicolumn{4}{|c|}{\textbf{Dependent Datasets}} \\ \hline
\multicolumn{4}{|c|}{\textit{Sentiment Classification (12 setups)}} \\ \hline 
AE-SCL-SR-MSDA & 10 & \textbf{6} & 10 \\ 
($\alpha = 0.05$)&  &  &  \\
\hline
AE-SCL-SR-MSDA & 6 & \textbf{2} & 8 \\ 
($\alpha = 0.01$)&  &   &  \\
\hline
\multicolumn{4}{|c|}{\textit{Word Similarity (12 datasets)}} \\ \hline
W2V-Glove & 8 & \textbf{6} & 7 \\ 
($\alpha = 0.05$)&  &  &  \\
\hline
W2V-Glove & 6 & \textbf{4} & 6 \\ 
($\alpha = 0.01$)&  &   &  \\
\hline
\end{tabular}
\caption{Replicability analysis results. The appropriate estimator for each scenario is in bold. For independent datasets $\alpha = 0.05$. $\hat{k}_{count}$ is based on the current practice in the NLP literature and does not have statistical guarantees regarding overestimation of the true $k$.
Likewise, $\hat{k}_{Fisher}$ does not provide  statistical guarantees regarding the overestimation of the true $k$ for dependent datasets.}
\label{table:khats}
\end{table}

\begin{table*}
\centering
  \begin{tabular}{| c | c | c | c | c | c | c | c | c |}
    \hline
    Model $|$ Data  & BC & BN & MZ & NW & PT & TC & WB \\ \hline
     Mate & 90.73  & 90.82  & 91.92  & 91.68 & 96.64  & 89.87  & 89.89  \\ \hline 
     SpaCy & 89.05  & 89.31& 89.29  & 89.52  & 95.27  & 87.65  & 87.40 \\ 
 $p-$val (Mate,SpaCy)  & ($10^{-4}$) & ($10^{-4}$) &  (0.0) & (0.0) &  ($2 \cdot 10^{-4}$) &  ($9 \cdot 10^{-4}$) & (0.0)\\
     \hline
     Redshift & 90.19 & 90.46 & 90.90  & 90.99  & 96.22 & 88.99  & 89.31  \\
 $p-$val (Mate,Redshift)  &  (0.0979) & (0.1662) & (0.0046) & (0.0376) & (0.0969) & (0.0912) &  (0.0823) \\
    \hline
    
  \end{tabular}
  \caption{UAS results for multi-domain dependency parsing. $p-$values are in parentheses. 
  } 
\label{table:DP}
\end{table*}

\begin{table}[t!]
\centering
  \begin{tabular}{| c | c | c | c |}
    \hline
    Language   & \textsc{Mimick} & Char$\rightarrow$Tag & $p-$value\\ \hline
    Kazakh     &        83.95 & 83.64 & 0.0944 \\ \hline
    Tamil$^{*}$      &  81.55 & 84.97  & 0.0001  \\ \hline
    Latvian    &  84.32 & 84.49  & 0.0623  \\ \hline
    Vietnamese &  84.22 & 84.85 &  0.0359  \\ \hline
    Hungarian$^{*}$  &  88.93 & 85.83 &  1.12e-08 \\ \hline
    Turkish    &  85.60 & 84.23 &  0.1461   \\ \hline
    Greek         & 93.63 & 94.05  &  0.0104   \\ \hline
    Bulgarian  &  93.16 & 93.03 &  0.1957   \\ \hline
    Swedish    &  92.30 & 92.27 &  0.0939   \\ \hline
    Basque$^{*}$         &  84.44 & 86.01 &  3.87e-10 \\ \hline
    Russian    &  89.72 & 88.65 &  0.0081   \\ \hline
    Danish     &  90.13 & 89.96 &  0.1016   \\ \hline
    Indonesian$^{*}$         & 89.34   & 89.81 & 0.0008 \\ \hline
    Chinese$^{*}$         & 85.69 & 81.84   & 0 \\ \hline
    Persian    &  93.58 & 93.53 &  0.4450  \\ \hline
    Hebrew     &  91.69 & 91.93 &  0.1025  \\ \hline
    Romanian         &  89.18 & 88.96 &  0.2198 \\ \hline
    English    &  88.45 & 88.89 &  0.0208  \\ \hline
    Arabic         & 90.58 & 90.49 & 0.0731 \\ \hline
    Hindi      &  87.77 & 87.92 &  0.0288  \\ \hline
    Italian    &  92.50 & 92.45 &  0.4812  \\ \hline
    Spanish    &  91.41 & 91.71 &  0.1176  \\ \hline
    Czech$^{*}$      &  90.81 & 90.17 &  2.91e-05\\ \hline
  \end{tabular}
  \caption{Multilingual POS tagging accuracy for the \textsc{Mimick} and the Char$\rightarrow$Tag models. $*$ indicates languages identified by the Holm procedure with $\alpha=0.05$ 
 .}
    \label{table:POS_multilingual}
\end{table}

\begin{table}[t!]
\centering
  \begin{tabular}{| c | c | c | c | c |}
    \hline
    Dataset & AE-SCL-SR & MSDA & $p-$value\\ \hline
    $B \rightarrow K$ & 0.8005 & 0.788 & 0.0268\\ \hline
    $B\rightarrow D^{*}$ & 0.8105 & 0.783 & 0.0011\\ \hline
    $B \rightarrow E$ & 0.7675 & 0.7455 & 0.0119\\ \hline
    $K \rightarrow B^{*}$ & 0.7295 & 0.7 & 0.0038\\ \hline
    $K \rightarrow D^{*,+}$ & 0.763 & 0.714 & 1.9e-06\\ \hline
    $K\rightarrow E$ & 0.84 & 0.824 & 0.018\\ \hline
	$D \rightarrow B$ & 0.773 & 0.7605 & 0.0186\\ \hline
    $D \rightarrow K^{*}$ & 0.8025 & 0.774 & 0.0014\\ \hline
    $D\rightarrow E^{*}$ & 0.781 & 0.75 & 0.0011\\ \hline
    $E \rightarrow B$ & 0.7115 & 0.7185 & 0.4823\\ \hline
    $E \rightarrow K$ & 0.8455 & 0.845 & 0.9507\\ \hline
    $E\rightarrow D^{*,+} $ & 0.745 & 0.71 & 0.0003\\ \hline
  \end{tabular}
  \caption{Cross-domain sentiment classification accuracy for models taken from \protect\cite{ziser2016neural}. In an $X\rightarrow Y$ setup, $X$ is the source domain and $Y$ is the target domain. $*$ and $+$ indicate domains identified by the Holm procedure with $\alpha=0.05$ and  $\alpha=0.01$, respectively.}
\label{table:domainsentiment_datasets} 
\end{table}

\begin{table}
\centering
  \begin{tabular}{| c | c | c | c | c |}
    \hline
    Dataset & W2V & GLOVE & $p-$val.\\ \hline
    WS353$^{*,+}$ & 0.7362 & 0.629  & $2e^{-5}$ \\ \hline
    WS353-SIM$^{*,+}$ & 0.7805 & 0.6979  & 0.0 \\ \hline
    WS353-REL & 0.6814 & 0.5706 & 0.2123 \\ \hline
    MC-30$^{*,+}$ & 0.8221 & 0.7773 & 0.0001 \\ \hline
    RG-65 & 0.8348 & 0.8117 & 0.3053 \\ \hline
    RW & 0.4819 & 0.4144 & 0.2426 \\ \hline
	MEN$^{*}$ & 0.796 & 0.7362 & 0.0021 \\ \hline
    MTurk-287 & 0.671 & 0.6475 & 0.2076 \\ \hline
    MTurk-771 & 0.7116 & 0.6842 & 0.0425 \\ \hline
    YP-130$^{*,+}$ & 0.504 & 0.5315 & 0.0 \\ \hline
    SimLex999$^{*}$ & 0.4621 & 0.3725 & 0.0015 \\ \hline
    $Verb-143$ & 0.4479 & 0.3275 & 0.0431 \\ \hline
  \end{tabular}
  \caption{Spearman's $\rho$ values for the best performing predict model (W2V-CBOW) of \protect\cite{Baroni:14} and the GLOVE model. $*$ and $+$ are as in Table~\ref{table:domainsentiment_datasets}.}
\label{table:wordsim} 
\end{table}

Table~\ref{table:khats} summarizes the replicability analysis results while Table~\ref{table:DP} -- \ref{table:wordsim} present task specific performance measures 
and $p-$values. 

\paragraph{Independent Datasets}
Dependency parsing (Tab. \ref{table:DP}) and multilingual POS tagging (Tab. \ref{table:POS_multilingual}) are our example tasks for this setup, where $\hat{k}_{Fisher}$ is our recommended valid estimator for
the number of cases where one algorithm outperforms another.

For dependency parsing, we compare two scenarios: (a) where in most domains the differences between the compared algorithms are quite large and the $p-$values are small (Mate vs. SpaCy); 
and (b) where in most domains the differences between the compared algorithms are smaller and the $p-$values are higher (Mate vs. Redshift). 
Our multilingual POS tagging scenario (\textsc{Mimick} vs. Char$\rightarrow$Tag) is more similar to scenario (b) in terms of the differences between the participating algorithms.

Table~\ref{table:khats} demonstrates the $\hat{k}$ estimators for the various tasks and scenarios. 
%
For dependency parsing, as expected, in scenario (a) where all the $p-$values are small, all estimators, even the error-prone $\hat{k}_{count}$, provide the same information.
In case (b) of dependency parsing, however, $\hat{k}_{Fisher}$ estimates the number of domains where Mate outperforms Redshift to be 5, while $\hat{k}_{count}$ estimates this number to be only 2. This is a substantial difference given that the total number of domains is 7. The $\hat{k}_{Bonferroni}$ estimator, that is valid under arbitrary dependencies and does not exploit the independence assumption as $\hat{k}_{Fisher}$ does, is even more conservative than $\hat{k}_{count}$ and its estimation is only 1. 

Perhaps not surprisingly, the multilingual POS tagging results are similar to case (b) of dependency parsing. Here, again, $\hat{k}_{count}$ is too conservative, estimating the number of languages with effect to be 11 (out of 23) while $\hat{k}_{Fisher}$ estimates this number to be 16 (an increase of 5/23 in the estimated number of languages with effect). $\hat{k}_{Bonferroni}$ is again more conservative, estimating the number of languages with effect to be only 6, which is not very surprising given that it does not exploit the independence between the datasets. These two examples of case (b) demonstrate that when the differences between the algorithms are quite small, $\hat{k}_{Fisher}$ may be more sensitive than the current practice in NLP for discovering the number of datasets with effect. 

To complete the analysis, we would like to name the datasets with effect. 
As discussed in Section~\ref{subsec:kestimator}, while this can be straightforwardly done by naming the datasets with the $\hat{k}$ smallest $p-$values, in general, this approach does not control the probability of identifying at least one dataset erroneously. We thus employ the Holm procedure for the identification task, noticing that  the number of datasets it identifies should be equal to 
the value of the $\hat{k}_{Bonferroni}$ estimator (Section~\ref{subsec:identification}).

Indeed, for dependency parsing in case (a), the Holm procedure identifies all seven domains as cases where Mate outperforms SpaCy,
while in case (b) it identifies only the MZ domain as a case where Mate outperforms Redshift. For multilingual POS tagging the Holm procedure identifies Tamil, Hungarian, Basque, Indonesian, Chinese and Czech as languages where \textsc{Mimick} outperforms Char$\rightarrow$Tag.
Expectedly, this analysis demonstrates that when the performance gap between two algorithms becomes narrower, inquiring for more information (i.e. identifying the domains with effect rather than just estimating their number), may come at the cost of weaker results.\footnote{For completeness, we also performed the analysis for the independent dataset setups with $\alpha = 0.01$. The results are ($\hat{k}_{count}$, $\hat{k}_{Bonferroni}$, $\hat{k}_{Fisher}$): Mate vs. Spacy: (7,7,7); Mate  vs. Redshift (1,0,2); \textsc{Mimick} vs. Char$\rightarrow$Tag: (7,5,13). The patterns are very similar to those discussed in the text.}

\paragraph{Dependent Datasets} In  cross-domain sentiment classification (Table~\ref{table:domainsentiment_datasets}) and word similarity prediction (Table~\ref{table:wordsim}), the involved datasets manifest mutual dependence. Particularly, each sentiment setup shares its test dataset with 2 other setups, while in word similarity WS-353 is the union of WS-353-REL and WS-353-SIM. As discussed in Section~\ref{sec:rep_analysis}, $\hat{k}_{Bonferroni}$ is the appropriate estimator of the number of cases one algorithm outperforms another.\roi{I am not sure if the sharing of training domains is also relevant here. I am afraid the reader will be confused.} 

The results in Table~\ref{table:khats} manifest the phenomenon demonstrated by the second toy example in Section~\ref{sec:toy}, which shows that when the datasets are dependent, $\hat{k}_{Fisher}$ as well as the error-prone $\hat{k}_{count}$ may be too optimistic regarding the number of datasets with effect. This stands in contrast to $\hat{k}_{Bonferroni}$ that controls the probability to overestimate the number of such datasets. 

Indeed, $\hat{k}_{Bonferroni}$ is much more conservative, yielding values of 6 ($\alpha$ = 0.05) and 2 ($\alpha$ = 0.01) for sentiment, and of 6 ($\alpha$ = 0.05) and 4 ($\alpha$ = 0.01) for word similarity. The differences from the conclusions that might have been drawn by $\hat{k}_{count}$ are again quite substantial. The difference between $\hat{k}_{Bonferroni}$  and $\hat{k}_{count}$ in sentiment classification is 4, which accounts to 1/3 of the 12 test setups. Even for word similarity, the difference between the two methods, which account to 2 for both $\alpha$ values, represents 1/6 of the 12 test setups.
The domains identified by the Holm procedure are marked in the tables.

\paragraph{Results Overview}

Our goal in this section is to demonstrate that the approach of simply looking at the number of datasets for which the difference between the performance of the algorithms reaches a predefined significance level, 
gives different results from our suggested statistically sound analysis. This approach is denoted here with $\hat{k}_{count}$ and shown to be statistically not valid in Sections \ref{subsec:adjmultiplicity} and \ref{sec:toy}.
We observe that this happens especially in evaluation setups where the differences between the algorithms are small for most datasets. In some cases, when the datasets are independent, our analysis has the power to declare a larger number of datasets with effect than the number of individual significant test values ($\hat{k}_{count}$). In other cases, when the datasets are interdependent, $\hat{k}_{count}$ is much too optimistic.

Our proposed analysis changes the observations that might have been made based on the  papers where the results analysed here were originally reported. For example, for the Mate-Redshift comparison (independent evaluation sets), we show that there is  evidence that the number of datasets with effect is much higher than one would assume based on counting the significant sets (5 vs. 2 out of 7 evaluation sets), giving a stronger claim regarding the superiority of Mate. In multingual POS tagging (again, an independent evaluation sets setup) our analysis shows evidence for 16 sets with effect compared to only 11 of the erroneous count method - a difference in 5 out of 23 evaluation sets (21.7\%). Finally, in the cross-domain sentiment classification and the word similarity judgment tasks (dependent evaluation sets), the unjustified counting method may be too optimistic (e.g. 10 vs. 6 out of 12 evaluation sets, for $\alpha = 0.05$ in the sentiment task), in favor of the new algorithms.

\section{Discussion and Future Directions}
\label{sec:conc}

We proposed a statistically sound replicability analysis framework for cases where algorithms are compared across multiple datasets. Our main contributions are: (a) analyzing the 
limitations of the current practice in NLP work: counting the datasets for which the difference between the algorithms reaches a predefined significance level; and 
(b) proposing a new framework that addresses both the estimation of the number of datasets with effect and the identification of such datasets.

The framework we propose addresses two different situations encountered in NLP: independent and dependent datasets. For dependent datasets, we assumed that the type of dependency cannot be determined. One could use more powerful methods if certain assumptions on the dependency between the test statistics could be made. For example, one could use the partial conjunction p-value based on Simes test for the global null hypothesis \cite{Simes:86}, which was proposed by \cite{benjamini2008screening} for the case where the test statistics satisfy certain positive dependency properties  (see Theorem 1 in \cite{benjamini2008screening}).  Using this partial conjunction p-value rather than the one based on Bonferroni, one may obtain higher values of $\hat{k}$ with the same statistical guarantee. Similarly, for the identification question, if certain positive dependency properties hold, Holm's procedure could be replaced by Hochberg's or Hommel's procedures \cite{Hochberg:88,Hommel:88} which are more powerful.

An alternative, more powerful multiple testing procedure for identification of datasets with effect, is the method in \newcite{benjamini1995controlling}, that controls the false discovery rate (FDR), a less strict error criterion than the one considered here. This method is more appropriate in cases where one may tolerate some errors as long as the proportion of errors among all the claims made is small, as expected to happen when the number of datasets grows.

We note that the increase in the number of evaluation datasets may have positive and negative aspects. As noted in Section \ref{sec:prev}, we believe that multiple comparisons are integral to NLP research when aiming to develop algorithms that perform well across languages and domains. On the other hand, experimenting with multiple  evaluation sets that reflect very similar linguistic phenomena may only complicate the comparison between alternative algorithms. 

In fact, our analysis is useful mostly where the datasets are heterogeneous, coming from different languages or domains. When they are just technically different but could potentially be just combined into a one big dataset, then we believe the question of \newcite{demvsar2006statistical}, whether at least one dataset shows evidence for effect,  is more appropriate. 

\section*{Acknowledgement}

The research of M. Bogomolov was supported by the Israel Science Foundation grant No. 1112/14. We thank Yuval Pinter for his great help with the multilingual experiments and for his useful feedback. We also thank Ruth Heller, Marten van Schijndel, Oren Tsur, Or Zuk and the ie@technion NLP group members for their useful comments. 

\bibliography{stat}

\begin{thebibliography}{}

\bibitem[\protect\citename{Agirre \bgroup et al.\egroup }2009]{agirre2009study}
Eneko Agirre, Enrique Alfonseca, Keith Hall, Jana Kravalova, Marius
  Pa{\c{s}}ca, and Aitor Soroa.
\newblock 2009.
\newblock A study on similarity and relatedness using distributional and
  wordnet-based approaches.
\newblock In {\em Proceedings of HLT-NAACL}.

\bibitem[\protect\citename{Al-Rfou \bgroup et al.\egroup }2013]{AlRfou:13}
Rami Al-Rfou, Bryan Perozzi, and Steven Skiena.
\newblock 2013.
\newblock Polyglot: Distributed word representations for multilingual nlp.
\newblock In {\em Proceedings of CoNLL}.

\bibitem[\protect\citename{Baker \bgroup et al.\egroup
  }2014]{baker2014unsupervised}
Simon Baker, Roi Reichart, and Anna Korhonen.
\newblock 2014.
\newblock An unsupervised model for instance level subcategorization
  acquisition.
\newblock In {\em Proceedings of EMNLP}.

\bibitem[\protect\citename{Baroni \bgroup et al.\egroup }2014]{Baroni:14}
Marco Baroni, Georgiana Dinu, and Germ{\'a}n Kruszewski.
\newblock 2014.
\newblock Don't count, predict! a systematic comparison of context-counting vs.
  context-predicting semantic vectors.
\newblock In {\em Proceedings of ACL}.

\bibitem[\protect\citename{Begley and Ellis}2012]{begley2012drug}
C.~Glenn Begley and Lee~M. Ellis.
\newblock 2012.
\newblock Drug development: Raise standards for preclinical cancer research.
\newblock {\em Nature}, 483(7391):531--533.

\bibitem[\protect\citename{Benjamini and Heller}2008]{benjamini2008screening}
Yoav Benjamini and Ruth Heller.
\newblock 2008.
\newblock Screening for partial conjunction hypotheses.
\newblock {\em Biometrics}, 64(4):1215--1222.

\bibitem[\protect\citename{Benjamini and
  Hochberg}1995]{benjamini1995controlling}
Yoav Benjamini and Yosef Hochberg.
\newblock 1995.
\newblock Controlling the false discovery rate: A practical and powerful
  approach to multiple testing.
\newblock {\em Journal of the royal statistical society. Series B
  (Methodological)}, pages 289--300.

\bibitem[\protect\citename{Benjamini \bgroup et al.\egroup
  }2006]{benjamini2006adaptive}
Yoav Benjamini, Abba~M. Krieger, and Daniel Yekutieli.
\newblock 2006.
\newblock Adaptive linear step-up procedures that control the false discovery
  rate.
\newblock {\em Biometrika}, pages 491--507.

\bibitem[\protect\citename{Benjamini \bgroup et al.\egroup
  }2009]{benjamini2009selective}
Yoav Benjamini, Ruth Heller, and Daniel Yekutieli.
\newblock 2009.
\newblock Selective inference in complex research.
\newblock {\em Philosophical Transactions of the Royal Society of London A:
  Mathematical, Physical and Engineering Sciences}, 367(1906):4255--4271.

\bibitem[\protect\citename{Berg-Kirkpatrick \bgroup et al.\egroup
  }2012]{berg2012empirical}
Taylor Berg-Kirkpatrick, David Burkett, and Dan Klein.
\newblock 2012.
\newblock An empirical investigation of statistical significance in nlp.
\newblock In {\em Proceedings of EMNLP-CoNLL}.

\bibitem[\protect\citename{Blitzer \bgroup et al.\egroup
  }2006]{blitzer2006domain}
John Blitzer, Ryan McDonald, and Fernando Pereira.
\newblock 2006.
\newblock Domain adaptation with structural correspondence learning.
\newblock In {\em Proceedings of EMNLP}.

\bibitem[\protect\citename{Blitzer \bgroup et al.\egroup
  }2007]{blitzer2007biographies}
John Blitzer, Mark Dredze, and Fernando Pereira.
\newblock 2007.
\newblock Biographies, bollywood, boom-boxes and blenders: Domain adaptation
  for sentiment classification.
\newblock In {\em Proceedings of ACL}.

\bibitem[\protect\citename{Bohnet}2010]{bohnet2010very}
Bernd Bohnet.
\newblock 2010.
\newblock Very high accuracy and fast dependency parsing is not a
  contradiction.
\newblock In {\em Proceedings of COLING}.

\bibitem[\protect\citename{Bruni \bgroup et al.\egroup
  }2012]{bruni2012distributional}
Elia Bruni, Gemma Boleda, Marco Baroni, and Nam-Khanh Tran.
\newblock 2012.
\newblock Distributional semantics in technicolor.
\newblock In {\em Proceedings of ACL}.

\bibitem[\protect\citename{Bruni \bgroup et al.\egroup }2014]{Bruni:14}
Elia Bruni, Nam-Khanh Tran, and Marco Baroni.
\newblock 2014.
\newblock Multimodal distributional semantics.
\newblock {\em Journal of Artificial Intelligence Research (JAIR)}, 49:1--47.

\bibitem[\protect\citename{Buchholz and Marsi}2006]{buchholz2006conll}
Sabine Buchholz and Erwin Marsi.
\newblock 2006.
\newblock Conll-x shared task on multilingual dependency parsing.
\newblock In {\em Proceedings of CoNLL}.

\bibitem[\protect\citename{Chan and Ng}2007]{chan2007domain}
Yee~Seng Chan and Hwee~Tou Ng.
\newblock 2007.
\newblock Domain adaptation with active learning for word sense disambiguation.
\newblock In {\em Proceedings of ACL}.

\bibitem[\protect\citename{Charniak}2000]{Charniak:00}
Eugene Charniak.
\newblock 2000.
\newblock A maximum-entropy-inspired parser.
\newblock In {\em Proceedings of HLT-NAACL}.

\bibitem[\protect\citename{Chen \bgroup et al.\egroup }2011]{chen2011automatic}
Minmin Chen, Yixin Chen, and Kilian~Q. Weinberger.
\newblock 2011.
\newblock Automatic feature decomposition for single view co-training.
\newblock In {\em Proceedings of ICML}.

\bibitem[\protect\citename{Choi \bgroup et al.\egroup }2015]{choi2015depends}
Jinho~D Choi, Joel Tetreault, and Amanda Stent.
\newblock 2015.
\newblock It depends: Dependency parser comparison using a web-based evaluation
  tool.
\newblock In {\em Proceedings of ACL}.

\bibitem[\protect\citename{Collaboration}2012]{open2012open}
Open~Science Collaboration.
\newblock 2012.
\newblock An open, large-scale, collaborative effort to estimate the
  reproducibility of psychological science.
\newblock {\em Perspectives on Psychological Science}, 7(6):657--660.

\bibitem[\protect\citename{Collins}2003]{Collins:03}
Michael Collins.
\newblock 2003.
\newblock Head-driven statistical models for natural language parsing.
\newblock {\em Computational linguistics}, 29(4):589--637.

\bibitem[\protect\citename{Daum{\'e}~III}2007]{daume2009frustratingly}
Hal Daum{\'e}~III.
\newblock 2007.
\newblock Frustratingly easy domain adaptation.
\newblock In {\em Proceedings of ACL}.

\bibitem[\protect\citename{De~Marneffe \bgroup et al.\egroup
  }2014]{DeMarneffe:14}
Marie-Catherine De~Marneffe, Timothy Dozat, Natalia Silveira, Katri Haverinen,
  Filip Ginter, Joakim Nivre, and Christopher~D. Manning.
\newblock 2014.
\newblock Stanford dependencies: A cross-linguistic typology.
\newblock In {\em Proceedings of LREC}.

\bibitem[\protect\citename{Dem{\v{s}}ar}2006]{demvsar2006statistical}
Janez Dem{\v{s}}ar.
\newblock 2006.
\newblock Statistical comparisons of classifiers over multiple data sets.
\newblock {\em Journal of Machine learning research}, 7(Jan):1--30.

\bibitem[\protect\citename{Dietterich}1998]{dietterich1998approximate}
Thomas~G. Dietterich.
\newblock 1998.
\newblock Approximate statistical tests for comparing supervised classification
  learning algorithms.
\newblock {\em Neural computation}, 10(7):1895--1923.

\bibitem[\protect\citename{Efron and Tibshirani}1994]{efron1994introduction}
Bradley Efron and Robert~J. Tibshirani.
\newblock 1994.
\newblock {\em An introduction to the bootstrap}.
\newblock CRC press.

\bibitem[\protect\citename{Farcomeni}2007]{farcomeni2007review}
Alessio Farcomeni.
\newblock 2007.
\newblock A review of modern multiple hypothesis testing, with particular
  attention to the false discovery proportion.
\newblock {\em Statistical Methods in Medical Research}.

\bibitem[\protect\citename{Faruqui and Dyer}2014]{faruqui-2014:SystemDemo}
Manaal Faruqui and Chris Dyer.
\newblock 2014.
\newblock Community evaluation and exchange of word vectors at wordvectors.org.
\newblock In {\em Proceedings of the ACL: System Demonstrations}.

\bibitem[\protect\citename{Finkelstein \bgroup et al.\egroup
  }2001a]{Finkelstein:01}
Lev Finkelstein, Evgeniy Gabrilovich, Yossi Matias, Ehud Rivlin, Zach Solan,
  Gadi Wolfman, and Eytan Ruppin.
\newblock 2001a.
\newblock Placing search in context: The concept revisited.
\newblock In {\em Proc. of WWW}.

\bibitem[\protect\citename{Finkelstein \bgroup et al.\egroup
  }2001b]{finkelstein2001placing}
Lev Finkelstein, Evgeniy Gabrilovich, Yossi Matias, Ehud Rivlin, Zach Solan,
  Gadi Wolfman, and Eytan Ruppin.
\newblock 2001b.
\newblock Placing search in context: The concept revisited.
\newblock In {\em Proceedings of WWW}.

\bibitem[\protect\citename{Guo \bgroup et al.\egroup }2009]{guo2009domain}
Honglei Guo, Huijia Zhu, Zhili Guo, Xiaoxun Zhang, Xian Wu, and Zhong Su.
\newblock 2009.
\newblock Domain adaptation with latent semantic association for named entity
  recognition.
\newblock In {\em Proceedings of HLT-NAACL}.

\bibitem[\protect\citename{Halawi \bgroup et al.\egroup }2012]{halawi2012large}
Guy Halawi, Gideon Dror, Evgeniy Gabrilovich, and Yehuda Koren.
\newblock 2012.
\newblock Large-scale learning of word relatedness with constraints.
\newblock In {\em Proceedings of ACM SIGKDD}.

\bibitem[\protect\citename{Heller \bgroup et al.\egroup
  }2014]{heller2014deciding}
Ruth Heller, Marina Bogomolov, and Yoav Benjamini.
\newblock 2014.
\newblock Deciding whether follow-up studies have replicated findings in a
  preliminary large-scale omics study.
\newblock {\em Proceedings of the National Academy of Sciences},
  111(46):16262--16267.

\bibitem[\protect\citename{Herndon \bgroup et al.\egroup
  }2014]{herndon2014does}
Thomas Herndon, Michael Ash, and Robert Pollin.
\newblock 2014.
\newblock Does high public debt consistently stifle economic growth? a critique
  of reinhart and rogoff.
\newblock {\em Cambridge journal of economics}, 38(2):257--279.

\bibitem[\protect\citename{Hill \bgroup et al.\egroup }2015]{Hill:15}
Felix Hill, Roi Reichart, and Anna Korhonen.
\newblock 2015.
\newblock Simlex-999: Evaluating semantic models with (genuine) similarity
  estimation.
\newblock {\em Computational Linguistics}, 41(4):665--695.

\bibitem[\protect\citename{Hill \bgroup et al.\egroup }2016]{hill2016simlex}
Felix Hill, Roi Reichart, and Anna Korhonen.
\newblock 2016.
\newblock Simlex-999: Evaluating semantic models with (genuine) similarity
  estimation.
\newblock {\em Computational Linguistics}.

\bibitem[\protect\citename{Hochberg}1988]{Hochberg:88}
Yosef Hochberg.
\newblock 1988.
\newblock A sharper bonferroni procedure for multiple tests of significance.
\newblock {\em Biometrika}, 75(4):800--802.

\bibitem[\protect\citename{Holm}1979]{Holm:79}
Sture Holm.
\newblock 1979.
\newblock A simple sequentially rejective multiple test procedure.
\newblock {\em Scandinavian Journal of Statistics}, 6(2):65--70.

\bibitem[\protect\citename{Hommel}1988]{Hommel:88}
Gerhard Hommel.
\newblock 1988.
\newblock A stagewise rejective multiple test procedure based on a modified
  bonferroni test.
\newblock {\em Biometrika}, 75(2):383--386.

\bibitem[\protect\citename{Honnibal and
  Johnson}2015]{honnibal-johnson:2015:EMNLP}
Matthew Honnibal and Mark Johnson.
\newblock 2015.
\newblock An improved non-monotonic transition system for dependency parsing.
\newblock In {\em Proceedings of EMNLP}.

\bibitem[\protect\citename{Honnibal \bgroup et al.\egroup
  }2013]{honnibal2013non}
Matthew Honnibal, Yoav Goldberg, and Mark Johnson.
\newblock 2013.
\newblock A non-monotonic arc-eager transition system for dependency parsing.
\newblock In {\em Proceedings of CoNLL}.

\bibitem[\protect\citename{Koehn and Schroeder}2007]{koehn2007experiments}
Philipp Koehn and Josh Schroeder.
\newblock 2007.
\newblock Experiments in domain adaptation for statistical machine translation.
\newblock In {\em Proceedings of the second workshop on statistical machine
  translation}.

\bibitem[\protect\citename{Koehn}2005]{koehn2005europarl}
Philipp Koehn.
\newblock 2005.
\newblock Europarl: A parallel corpus for statistical machine translation.
\newblock In {\em Proceedings of the tenth Machine Translation Summit}.

\bibitem[\protect\citename{Leek and Peng}2015]{Leek:15}
Jeffrey~T. Leek and Roger~D Peng.
\newblock 2015.
\newblock Opinion: Reproducible research can still be wrong: Adopting a
  prevention approach.
\newblock {\em Proceedings of the National Academy of Sciences},
  112(6):1645--1646.

\bibitem[\protect\citename{Levy and Goldberg}2014]{Levy:14}
Omer Levy and Yoav Goldberg.
\newblock 2014.
\newblock Dependency-based word embeddings.
\newblock In {\em Proceedings of ACL}.

\bibitem[\protect\citename{Ling \bgroup et al.\egroup }2015]{Ling:15}
Wang Ling, Chris Dyer, Alan~W. Black, Isabel Trancoso, Ramon Fermandez, Silvio
  Amir, Luis Marujo, and Tiago Luis.
\newblock 2015.
\newblock Finding function in form: Compositional character models for open
  vocabulary word representation.
\newblock In {\em Proceedings of EMNLP}.

\bibitem[\protect\citename{Loughin}2004]{loughin2004systematic}
Thomas~M. Loughin.
\newblock 2004.
\newblock A systematic comparison of methods for combining p-values from
  independent tests.
\newblock {\em Computational statistics \& data analysis}, 47(3):467--485.

\bibitem[\protect\citename{Luong \bgroup et al.\egroup
  }2013]{Luong-etal:conll13:morpho}
Minh-Thang Luong, Richard Socher, and Christopher~D. Manning.
\newblock 2013.
\newblock Better word representations with recursive neural networks for
  morphology.
\newblock In {\em Proceedings of CoNLL}.

\bibitem[\protect\citename{Marcus \bgroup et al.\egroup }1993]{Marcus:93}
Mitchell~P. Marcus, Mary~Ann Marcinkiewicz, and Beatrice Santorini.
\newblock 1993.
\newblock Building a large annotated corpus of english: The penn treebank.
\newblock {\em Computational linguistics}, 19(2):313--330.

\bibitem[\protect\citename{Marrese-Taylor and Matsuo}2017]{Marrese-Taylor:17}
Edison Marrese-Taylor and Yutaka Matsuo.
\newblock 2017.
\newblock Replication issues in syntax-based aspect extraction for opinion
  mining.
\newblock In {\em Proceedings of the Student Research Workshop at EACL}.

\bibitem[\protect\citename{McNemar}1947]{mcnemar1947note}
Quinn McNemar.
\newblock 1947.
\newblock Note on the sampling error of the difference between correlated
  proportions or percentages.
\newblock {\em Psychometrika}, 12(2):153--157.

\bibitem[\protect\citename{Mikolov \bgroup et al.\egroup }2013]{Mikolov:13}
Tomas Mikolov, Ilya Sutskever, Kai Chen, Gregory~S. Corrado, and Jeffrey Dean.
\newblock 2013.
\newblock Distributed representations of words and phrases and their
  compositionality.
\newblock In {\em Proceedings of NIPS}.

\bibitem[\protect\citename{Miller and Charles}1991]{miller1991contextual}
George~A. Miller and Walter~G. Charles.
\newblock 1991.
\newblock Contextual correlates of semantic similarity.
\newblock {\em Language and cognitive processes}, 6(1):1--28.

\bibitem[\protect\citename{Moonesinghe \bgroup et al.\egroup
  }2007]{moonesinghe2007most}
Ramal Moonesinghe, Muin~J. Khoury, and A.~Cecile J.~W. Janssens.
\newblock 2007.
\newblock Most published research findings are false—but a little replication
  goes a long way.
\newblock {\em PLoS Med}, 4(2):e28.

\bibitem[\protect\citename{N{\'e}v{\'e}ol \bgroup et al.\egroup
  }2016]{Neveol:16}
Aur{\'e}lie N{\'e}v{\'e}ol, Cyril Grouin, Kevin~Bretonnel Cohen, and Aude
  Robert.
\newblock 2016.
\newblock Replicability of research in biomedical natural language processing:
  a pilot evaluation for a coding task.
\newblock {\em Proceedings of EMNLP}.

\bibitem[\protect\citename{Nilsson \bgroup et al.\egroup
  }2007]{nilsson2007conll}
Jens Nilsson, Sebastian Riedel, and Deniz Yuret.
\newblock 2007.
\newblock The conll 2007 shared task on dependency parsing.
\newblock In {\em Proceedings of CoNLL}.

\bibitem[\protect\citename{{\'O S\'eaghdha} and
  Korhonen}2014]{OSeaghdha:Korhonen:14}
Diarmuid {\'O S\'eaghdha} and Anna Korhonen.
\newblock 2014.
\newblock Probabilistic distributional semantics.
\newblock {\em Computational Linguistics}, 40(3):587--631.

\bibitem[\protect\citename{Patil \bgroup et al.\egroup
  }2016]{patil2016statistical}
Prasad Patil, Roger~D. Peng, and Jeffrey Leek.
\newblock 2016.
\newblock A statistical definition for reproducibility and replicability.
\newblock {\em bioRxiv}.

\bibitem[\protect\citename{Peng}2011]{Peng:11}
Roger~D. Peng.
\newblock 2011.
\newblock Reproducible research in computational science.
\newblock {\em Science}, 334(6060):1226--1227.

\bibitem[\protect\citename{Pennington \bgroup et al.\egroup
  }2014]{Pennington:14}
Jeffrey Pennington, Richard Socher, and Christopher Manning.
\newblock 2014.
\newblock {GloVe: G}lobal vectors for word representation.
\newblock In {\em Proceedings of EMNLP}.

\bibitem[\protect\citename{Petrov and McDonald}2012]{petrov2012overview}
Slav Petrov and Ryan McDonald.
\newblock 2012.
\newblock Overview of the 2012 shared task on parsing the web.
\newblock In {\em Notes of the First Workshop on Syntactic Analysis of
  Non-Canonical Language (SANCL)}.

\bibitem[\protect\citename{Pinter \bgroup et al.\egroup }2017]{Pinter:17}
Yuval Pinter, Robert Guthrie, and Jacob Eisenstein.
\newblock 2017.
\newblock Mimicking word embeddings using subword rnns.
\newblock In {\em Proceedings of EMNLP}.

\bibitem[\protect\citename{Pradhan \bgroup et al.\egroup
  }2013]{pradhan2013towards}
Sameer Pradhan, Alessandro Moschitti, Nianwen Xue, Hwee~Tou Ng, Anders
  Bj{\"o}rkelund, Olga Uryupina, Yuchen Zhang, and Zhi Zhong.
\newblock 2013.
\newblock Towards robust linguistic analysis using ontonotes.
\newblock In {\em Proceedings of CoNLL}.

\bibitem[\protect\citename{Radinsky \bgroup et al.\egroup
  }2011]{radinsky2011word}
Kira Radinsky, Eugene Agichtein, Evgeniy Gabrilovich, and Shaul Markovitch.
\newblock 2011.
\newblock A word at a time: Computing word relatedness using temporal semantic
  analysis.
\newblock In {\em Proceedings of WWW}.

\bibitem[\protect\citename{Rubenstein and
  Goodenough}1965]{rubenstein1965contextual}
Herbert Rubenstein and John~B. Goodenough.
\newblock 1965.
\newblock Contextual correlates of synonymy.
\newblock {\em Communications of the ACM}, 8(10):627--633.

\bibitem[\protect\citename{Schwartz \bgroup et al.\egroup
  }2015]{schwartz-reichart-rappoport:2015:Conll}
Roy Schwartz, Roi Reichart, and Ari Rappoport.
\newblock 2015.
\newblock Symmetric pattern based word embeddings for improved word similarity
  prediction.
\newblock In {\em Proceedings of CoNLL}.

\bibitem[\protect\citename{Silberer and Lapata}2014]{Silberer:14}
Carina Silberer and Mirella Lapata.
\newblock 2014.
\newblock Learning grounded meaning representations with autoencoders.
\newblock In {\em Proceedings of ACL}.

\bibitem[\protect\citename{Simes}1986]{Simes:86}
R~John Simes.
\newblock 1986.
\newblock An improved bonferroni procedure for multiple tests of significance.
\newblock {\em Biometrika}, pages 751--754.

\bibitem[\protect\citename{Snow \bgroup et al.\egroup }2008]{snow2008cheap}
Rion Snow, Brendan O'Connor, Daniel Jurafsky, and Andrew~Y. Ng.
\newblock 2008.
\newblock Cheap and fast---but is it good?: Evaluating non-expert annotations
  for natural language tasks.
\newblock In {\em Proceedings of EMNLP}.

\bibitem[\protect\citename{S{\o}gaard \bgroup et al.\egroup
  }2014]{sogaard2014s}
Anders S{\o}gaard, Anders Johannsen, Barbara Plank, Dirk Hovy, and
  H{\'e}ctor~Mart{\'\i}nez Alonso.
\newblock 2014.
\newblock What's in a p-value in nlp?
\newblock In {\em Proceedings of CoNLL}.

\bibitem[\protect\citename{S{\o}gaard}2013]{sogaard2013estimating}
Anders S{\o}gaard.
\newblock 2013.
\newblock Estimating effect size across datasets.
\newblock In {\em Proceedings of HLT-NAACL}.

\bibitem[\protect\citename{Steiger}1980]{steiger1980tests}
James~H. Steiger.
\newblock 1980.
\newblock Tests for comparing elements of a correlation matrix.
\newblock {\em Psychological bulletin}, 87(2):245--251.

\bibitem[\protect\citename{Weischedel \bgroup et al.\egroup
  }2011]{weischedel2011ontonotes}
Ralph Weischedel, Eduard Hovy, Mitchell Marcus, Martha Palmer, Robert Belvin,
  Sameer Pradhan, Lance Ramshaw, and Nianwen Xue.
\newblock 2011.
\newblock Ontonotes: A large training corpus for enhanced processing.
\newblock {\em Handbook of Natural Language Processing and Machine Translation.
  Springer}.

\bibitem[\protect\citename{Wilcoxon}1945]{Wilcoxon:45}
Frank Wilcoxon.
\newblock 1945.
\newblock Individual comparisons by ranking methods.
\newblock {\em Biometrics bulletin}, 1(6):80--83.

\bibitem[\protect\citename{Yang and Powers}]{yang2006verb}
Dongqiang Yang and David~MW. Powers.
\newblock Verb similarity on the taxonomy of wordnet.
\newblock In {\em Proceedings of the 3rd International WordNet Conference}.

\bibitem[\protect\citename{Yeh}2000]{yeh2000more}
Alexander Yeh.
\newblock 2000.
\newblock More accurate tests for the statistical significance of result
  differences.
\newblock In {\em Proceedings of CoNLL}.

\bibitem[\protect\citename{Ziser and Reichart}2017]{ziser2016neural}
Yftah Ziser and Roi Reichart.
\newblock 2017.
\newblock Neural structural correspondence learning for domain adaptation.
\newblock In {\em Proceedings of CoNLL}.

\end{thebibliography}
\bibliographystyle{acl2012}
\end{document}